\documentclass{article}

\usepackage[hidelinks]{hyperref}
\usepackage{spconf, amsmath, bm, graphicx, cite, siunitx, booktabs, listings, inconsolata, multicol, enumitem, float}
\usepackage[acronym]{glossaries}
\usepackage[capitalize]{cleveref}
\usepackage{tikz}
\usepackage{doi}

\newacronym{2d}{2D}{2-dimensional}
\newacronym{3d}{3D}{3-dimensional}
\newacronym{adam}{Adam}{Adaptive Moment Estimation}
\newacronym{ai}{AI}{Artificial Intelligence}
\newacronym{ann}{ANN}{Artificial Neural Network}
\newacronym{cnn}{CNN}{Convolutional Neural Network}
\newacronym{dl}{DL}{Deep Learning}
\newacronym{dnn}{DNN}{Deep Neural Network}
\newacronym{es}{\textit{es}}{exponent size}
\newacronym{fc}{FC}{Fully Connected}
\newacronym{fpu}{FPU}{Floating-Point Unit}
\newacronym{ilsvrc}{ILSVRC}{ImageNet Large Scale Visual Recognition Challenge}
\newacronym{msc}{MSc}{Master of Science}
\newacronym{mse}{MSE}{Mean Squared Error}
\newacronym{nbits}{\textit{nbits}}{number of bits}
\newacronym{nag}{NAG}{Nesterov Accelerated Gradient}
\newacronym{nlp}{NLP}{Natural Language Processing}
\newacronym{nn}{NN}{Neural Network}
\newacronym{ppu}{PPU}{Posit Processing Unit}
\newacronym{relu}{ReLU}{Rectified Linear Unit}
\newacronym{rmsprop}{RMSprop}{Root Mean Square Propagation}
\newacronym{sgd}{SGD}{Stochastic Gradient Descent}
\newacronym{tanh}{TanH}{Hyperbolic Tangent}
\newacronym{vgg}{VGG}{Visual Geometry Group}

\newacronym{asic}{ASIC}{Application Specific Integrated Circuits}
\newacronym{cpu}{CPU}{Central Processing Unit}
\newacronym{fpga}{FPGA}{Field Programmable Gate Arrays}
\newacronym{fcnn}{FCNN}{Fully Connected Neural Network}
\newacronym{fp}{FP}{floating-point}
\newacronym{gpu}{GPU}{Graphics Processing Unit}
\newacronym{ieee754}{IEEE 754}{IEEE Standard for Floating-Point Arithmetic}
\newacronym{sp}{SP}{single-precision}

\title{POSITNN: TRAINING DEEP NEURAL NETWORKS WITH MIXED LOW-PRECISION POSIT}
\name{Gonçalo Raposo \qquad Pedro Tomás \qquad Nuno Roma }
\address{INESC-ID, Instituto Superior Técnico, Universidade de Lisboa, Portugal}
\newcommand\mynotice[1]{
	\begin{tikzpicture}[remember picture,overlay]
	\node[anchor=north,yshift=-10pt] at (current page.north) {\parbox{\dimexpr\textwidth-\fboxsep-\fboxrule\relax}{#1}};
	\end{tikzpicture}
}

\newcommand\mycopyrightnotice[1]{
	\begin{tikzpicture}[remember picture,overlay]
	\node[anchor=south,yshift=10pt] at (current page.south) {\parbox{\dimexpr\textwidth-\fboxsep-\fboxrule\relax}{#1}};
	\end{tikzpicture}
}

\begin{document}
	%
	\maketitle
	\mynotice{\centering\small Published in ICASSP 2021 - 2021 IEEE International Conference on Acoustics, Speech and Signal Processing (ICASSP). \doi{10.1109/ICASSP39728.2021.9413919}}
	\mycopyrightnotice{
		\footnotesize \textcopyright~2021 IEEE. Personal use of this material is permitted. Permission from IEEE must be obtained for all other uses, in any current or future media, including reprinting/republishing this material for advertising or promotional purposes, creating new collective works, for resale or redistribution to servers or lists, or reuse of any copyrighted component of this work in other works.}
	\begin{abstract}
		Low-precision formats have proven to be an efficient way to reduce not only the memory footprint but also the hardware resources and power consumption of deep learning computations.  Under this premise, the posit numerical format appears to be a highly viable substitute for the IEEE floating-point, but its application to neural networks training still requires further research. Some preliminary results have shown that 8-bit (and even smaller) posits may be used for inference and 16-bit for training, while maintaining the model accuracy. The presented research aims to evaluate the feasibility to train deep convolutional neural networks using posits. For such purpose, a software framework was developed to use simulated posits and quires in end-to-end training and inference. This implementation allows using any bit size, configuration, and even mixed precision, suitable for different precision requirements in various stages.
		The obtained results suggest that 8-bit posits can substitute 32-bit floats during training with no negative impact on the resulting loss and accuracy.
	\end{abstract}
	\begin{keywords}
		Posit numerical format, low-precision arithmetic, deep neural networks, training, inference
	\end{keywords}
	%
	\section{Introduction}
	\label{sec:intro}
	
	\gls{dl} is, nowadays, one of the hottest topics in signal processing research, spanning across multiple applications. This is a highly demanding computational field, since, in many cases, better performance and generality result in increased complexity and deeper models \cite{Thompson2020}. For example, the recently published language model GPT-3, the largest ever trained network with 175 billion parameters, would require 355 years and \$4.6M to train on a Tesla V100 cloud instance\cite{Li2020}. Therefore, it is increasingly important to optimize the energy consumption required by the training process. Although algorithmic approaches may contribute to these goals, computing architectures advances are also fundamental \cite{Schmidhuber2015}.
	
	The computations involved in \gls{dl} mostly use the \glsunset{ieee754}\acrshort{ieee754} \gls{sp} \gls{fp} format \cite{IEEE2019}, with 32 bits. However, recent research has achieved comparable precision with smaller numerical formats. The novel posit format \cite{Gustafson2017}, designed as a direct drop-in replacement for float (i.e., IEEE SP FP), provides a wider dynamic range, higher accuracy, and simpler hardware. Moreover, each posit format has a corresponding exact accumulator, named quire, which is particularly useful for the frequent dot products in \gls{dl}.
	
	Contrasting with the \acrshort{ieee754} FP, the posit numerical format may be used with any size and has been shown to be able to provide more accurate operations than floats, while using fewer bits. Posits may even use sizes that are not multiples of 8, which could be exploited in \gls{fpga} or \gls{asic} to obtain optimal efficiency and performance.
	
	
	
	However, most published studies regarding the application of the posit format to \glspl{dnn} rely on the inference stage \cite{Cococcioni2018, Johnson2018, Langroudi2018, Carmichael2019, Carmichael2019a, Langroudi2019, Langroudi2020}. The models are trained using floats and are later quantized to posits to be used for inference. Nevertheless, the inference phase tends to be less sensitive to errors than the training phase, making it easier to achieve good performance using \{5..8\}-bit posits.
	
	In contrast, exploiting the use of posits during the training phase is a more compelling topic since this is the most computationally demanding stage. The first time posits were used in this context was in \cite{Montero2019}, by training a \gls{fcnn} for a binary classification problem using \{8, 10, 12, 16, 32\}-bit posits. Later, in \cite{Langroudi2019b, Langroudi2019a}, a \gls{fcnn} was trained for MNIST and Fashion MNIST using \{16, 32\}-bit posits. In \cite{Lu2019, Lu2020}, \glspl{cnn} were trained using a mix of \{8, 16\}-bit posits, but still relying on floats for the first epoch and layer computations. More recently, in \cite{Murillo2020}, a \gls{cnn} was trained for CIFAR-10 but using only \{16, 32\}-bit posits.
	
	
	
	
	Under the premise of these previous works, the research that is now presented goes a step further by extending the implementation of \glspl{dnn} in a more general and feature-rich approach. Hence, the original contributions of this paper are:
	
	\begin{itemize}[leftmargin=*]
		\item \textbf{open-source framework}\footnote{\href{https://github.com/hpc-ulisboa/posit-neuralnet}{Available at: https://github.com/hpc-ulisboa/posit-neuralnet}} to natively perform inference and training with posits of any precision (number of bits and exponent size) and quires; it was developed in C++ and adopts a similar API as PyTorch, with multithread support;
		
		\item adaptation of the framework to support \textbf{mixed-precision}, with different stages (forward, backward, gradient, optimizer, and loss) operating under different posit formats;
		
		
		
		\item training \glspl{cnn} with only \textbf{8 to 12-bit posits} without impacting on the achieved model accuracy.
	\end{itemize}
	
	
	\section{Posit numbering system}
	
	Among the several different numbering formats that have been proposed to represent real numbers~\cite{Sousa2020}, the \gls{ieee754} single-precision floating-point (float) is the most widely adopted. It decomposes a number into a sign (1-bit), exponent (8-bits) and mantissa (23-bits):
	\begin{equation}
	f = \left(-1\right)^\text{sign} \times \text{2}^{\text{exponent}-127} \times \text{mantissa}.
	\label{eq:float}
	\end{equation}
	However, it has also been observed that many application domains do not need nor make use of the total accuracy and wide dynamic range that is made available by \gls{ieee754}, often compromising the resulting system optimization in terms of hardware resources, performance, and energy efficiency. One of such domains is \gls{dnn} training, where most of the computations are zero-centered.
	
	To overcome these issues, the Posit numbering system~\cite{Gustafson2017} was recently proposed as a new alternative to \gls{ieee754}. Posit is characterized by a fixed size/\gls{nbits} and an \gls{es}, being composed by the following fields: sign (1-bit), regime (variable bits), exponent (0..\gls{es}-bits), and fraction (remaining bits)~\cite{Group2018}. It is decoded as in \cref{eq:posit}.
	\begin{equation}
	p = \left(-1\right)^\text{sign} \times 2^{2^\textit{es} \times k} \times 2^\text{exponent} \times \left( 1 + \text{fraction} \right).
	\label{eq:posit}
	\end{equation}
	
	When the number is negative, the two's complement has to be applied before decoding the other fields. The regime bits are decoded by measuring $k$, determined by their run-length.
	
	
	A particular characteristic of Posit, and perhaps the most interesting aspect for \acrshort{dnn} applications, refers to the distribution of its values, resembling a log-normal distribution (see \cref{fig:posit_distribution}), which is similar to the normal distribution of the values commonly found in \glspl{dnn}.
	%
	%
	Another interesting point is the definition of the quire, a Kulisch-like large accumulator~\cite{Kulisch2012} designed to contain exact sums of products of posits. \cref{tab:posit_quire} shows the recommended posit and quire configurations.

	\section{Deep learning posit framework}
	
	Current \gls{dnn} frameworks (such as PyTorch and TensorFlow/Keras) do not natively support the posit data type. As a result, the whole set of functions and operators would need to be reimplemented, in order to take advantage of this new numbering system. As such, it was decided to develop an entirely new framework, from scratch, in order to ensure better control of its inner operations and exploit them for the posit data format.
	
	\subsection{PositNN Framework}
	
	The developed framework, named PositNN, was based on the PyTorch API for C++ (LibTorch), thus inheriting its program functions and data flow. As a result, any user familiar with PyTorch may easily port their networks and models to PositNN. As an example, a comparison between PyTorch and the proposed framework regarding the declaration of a 1-layer model is shown in \cref{fig:framework} (left and center). The overall structure and functions are very similar, the only difference being the declaration of the backward function, since the proposed framework does not currently support automatic differentiation.
	
\begin{figure}[htb]
		\centering
		\includegraphics[width=.9\columnwidth]{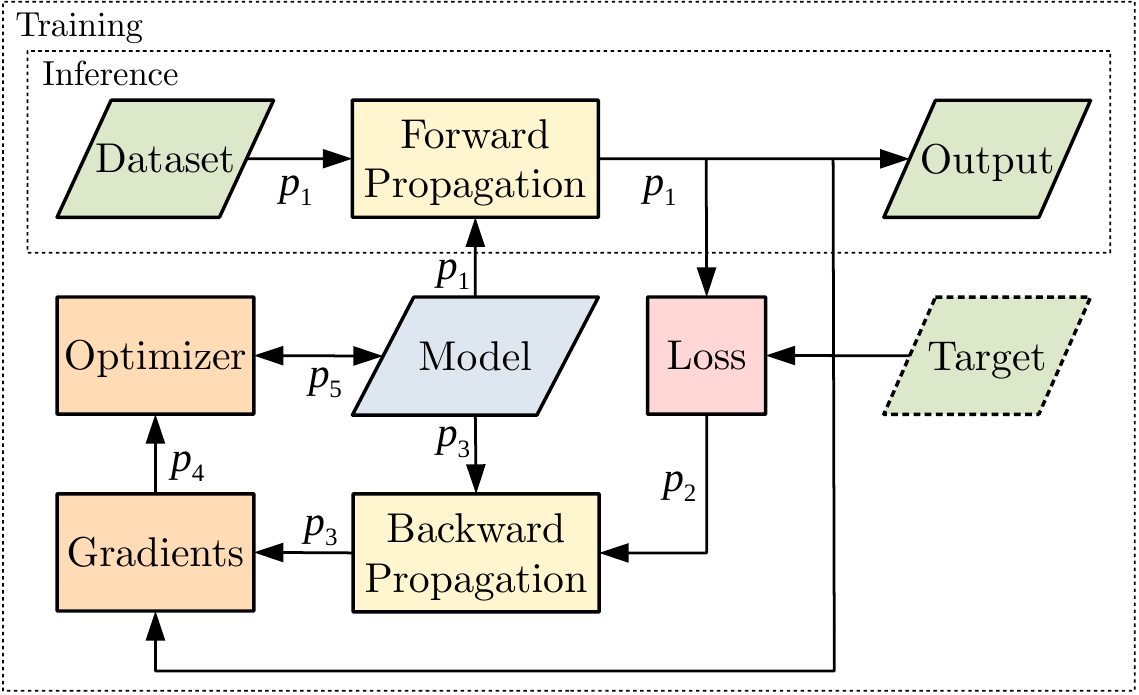}
		\vspace*{-0.5\baselineskip}
		\caption{\gls{dnn} training diagram, starting at the dataset. The various $p_i$, with $i=\{1..5\}$, represent the different posit precisions that may be used throughout the proposed framework.}
		\vspace*{-0.5\baselineskip}
		\label{fig:diagram}
	\end{figure}

	Despite being compared against a full-fledged framework, like PyTorch, the proposed framework is also capable of performing \gls{dnn} inference and training with the most common models and functions. A complete list of the supported functionalities is shown in \cref{fig:framework} (right), which allow implementing all the stages illustrated in \cref{fig:diagram}. Thus, common \glspl{cnn}, such as LeNet-5, CifarNet, AlexNet, and others, are fully supported. Moreover, the framework allows the user to extend it with custom functions or combine it with existing ones (e.g.\ from PyTorch).
	
	\begin{table}[t]
		\vspace*{-0.5\baselineskip}
		\centering
		\caption{Main properties of posit formats according to \cite{Group2018}.}
		\label{tab:posit_quire}
		\begin{tabular}{@{}lcccc@{}}
			\toprule
			nbits & $8$ & $16$ & $32$ & $64$ \\ \midrule
			es & $0$ & $1$ & $2$ & $3$ \\
			dynamic range & $2^{\pm 6}$ & $2^{\pm 28}$ & $2^{\pm 120}$ & $2^{\pm 496}$ \\
			quire bits & $32$ & $128$ & $512$ & $2048$ \\
			dot product limit & $127$ & $32767$ & $2^{31}-1$ & $2^{63}-1$ \\ \bottomrule
		\end{tabular}
	\end{table}        
	\begin{figure}[t]
		\centering
		\includegraphics[width=1\columnwidth]{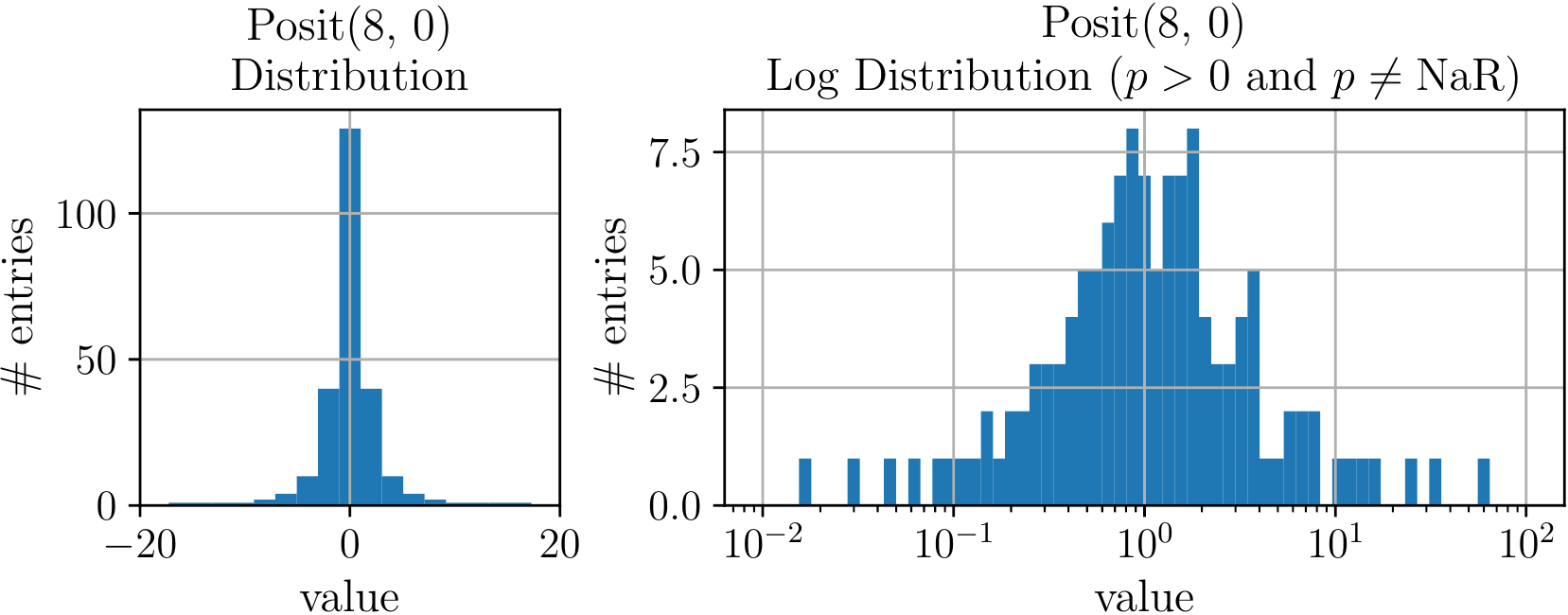}
		\vspace*{-1.5\baselineskip}
		\caption{Distribution of posit(8, 0) in linear and log scale.}
		\vspace*{-0.5\baselineskip}
		\label{fig:posit_distribution}
	\end{figure}
	
	\begin{figure*}[htb]
		\begin{minipage}[b]{.69\textwidth}
			{\input{code.tex}}
		\end{minipage}%
		\hfill
		\fbox{\begin{minipage}[b]{.28\textwidth}
				\footnotesize
				\begin{itemize}[leftmargin=*, itemindent=-2ex, itemsep=0.5em]
					\item {\bf Activation functions:}\\ ReLU, Sigmoid, TanH
					\item {\bf Layers:}\\  Linear, Convolutional, \\ Average and Maximum Pooling,\\ Batch Normalization, Dropout
					\item {\bf Loss functions:}\\ Cross Entropy,\\ \gls{mse}
					\item {\bf Optimizer:}\\ \gls{sgd}
					\item {\bf Utilities:}\\ StdTensor, convert PyTorch tensors, mixed precision tensor, save and load model, scaled gradients
				\end{itemize}
		\end{minipage}}
		\caption{Comparison of PyTorch (left) and the proposed framework (center). Implemented functionalities of PositNN (right).}
		\vspace*{-1\baselineskip}
		\label{fig:framework}
	\end{figure*}

	\subsection{Posit variables}
	
	Among the several libraries already available to implement posit operators in software \cite{NGATeam2019}, the Universal\footnote{Available at: \href{https://github.com/stillwater-sc/universal}{https://github.com/stillwater-sc/universal}} library was selected, thanks to its comprehensive support for any posit configuration and quires. Furthermore, C++ classes and function templates are generically used to implement different posits. Therefore, declaring a posit(8, 0) variable \texttt{p} equal to 1 is as simple as:
	\begin{lstlisting}[language=C++, basicstyle=\small\ttfamily, frame=leftline]
#include <universal/posit/posit>
sw::unum::posit<8, 0> p = 1;
	\end{lstlisting}
	Moreover, all the main operations specified in the Posit Standard \cite{Group2018} are fully supported and implemented. Furthermore, the proposed framework adopts bitwise operations whenever possible, thus avoiding operating with intermediate float representations, since this could introduce errors regarding a native implementation.
	
	
	\subsection{Implementation}
	
	Posit tensors are stored as StdTensors, a class implemented using only the C++ Standard Library. Data is internally stored in a one-dimensional dynamic vector and the multidimensional strides are automatically accounted for.
	
	Given that some stages are more sensitive to numerical errors, the proposed framework supports different precisions per stage, as depicted in the arrows of \cref{fig:diagram}. Although not illustrated, it even allows the model to use different precisions per layer. To accomplish that, the weights are stored in a class where members are copies with different posit configurations. Hence, each layer and stage converts its posit tensors to the appropriate precisions and seamlessly updates the copies after every change. It also has the option to use quires for the accumulations, significantly improving the accuracy of matrix multiplications, convolutions, etc.
	
	In order to take the maximum advantage of the \acrshort{cpu}, most functions were conveniently parallelized and implemented with multithread support, thus dividing each mini-batch by different workers. In matrix multiplication, this corresponds to splitting the left operand by rows, performing the computation, and then concatenating the results. The threads were implemented using std::thread.
	
	The proposed framework could also be adapted to support other data types, since most functions are independent of the posit format, except those that use the quire to accumulate.
	
	
	
	\vspace{-0.5\baselineskip}
	\section{Experimental Evaluation}
	\vspace{-0.5\baselineskip}
	
	By making use of the developed framework, the presented research started by studying how much can the posit precision be decreased without penalizing the \gls{dnn} model accuracy. Then, the best configuration was used to train a deeper model on a more complex dataset. In this evaluation, small accuracy differences ($<1\%$) were assumed to be caused solely by the randomness of the training process and not exactly by lack of precision of the numerical format.
	
	For the initial evaluation, the 5-layer \gls{cnn} LeNet-5 was trained on Fashion MNIST (a more complex dataset than the ordinary MNIST) during 10 epochs. Just as in \cite{Langroudi2019a, Murillo2020}, posit(16, 1) was first used everywhere and decreased until posit(8, 0) (see \cref{tab:same_posit_without_quire}).
	\begin{table}[b]
		\vspace*{-\baselineskip}
		\centering
		\caption{Accuracy of LeNet-5 trained on Fashion MNIST using posit and without quire, using float for reference.}
		\label{tab:same_posit_without_quire}
		\begin{tabular}{@{}l|c|cccc@{}}
			\toprule
			Posit & \textbf{Float} & $(16, 1)$ & $(12, 1)$ & $(10, 1)$ & $(8, 0)$  \\ \midrule
			Accuracy [\%] & \textbf{\num{90.42}} & \num{90.87} & \num{90.15} & \num{88.15} & \num{10.00} \\ \bottomrule
		\end{tabular}
	\end{table}
	
	As expected, posit(16, 1) achieves a float-like accuracy, and narrower precisions, such as posit(12, 1) and posit(10, 1), are also usable for training, the latter incurring in some accuracy loss. However, when trained using posit(8, 0), the model accuracy does not improve and fixes at $\SI{10}{\percent}$ (equivalent to randomly classifying a 10-class dataset), probably due to the narrow dynamic range (as seen in \cref{fig:posit_distribution}). This hypothesis was subsequently evaluated by using a different exponent size (\gls{es}) and using quires for the accumulations (see \cref{tab:es_quire}). The obtained results confirmed the hypothesis, showing that the precision of the 8-bit model slightly increases when using quires, especially when the posit exponent size is $\textit{es}=2$.
	\begin{table}[t]
		\centering
		\caption{Accuracy of LeNet-5 trained on Fashion MNIST using posit and quire. Posit8 is tested with different \gls{es}.}
		\label{tab:es_quire}
		\resizebox{\columnwidth}{!}{\begin{tabular}{@{}l|c|cccc@{}}
				\toprule
				Posit with quire & \textbf{Float} & $(10, 1)$ & $(8, 0)$ & $(8, 1)$ & $(8, 2)$ \\ \midrule
				Accuracy [\%] & \textbf{\num{90.42}} & \num{88.40} & \num{13.84} & \num{12.86} & \num{19.39} \\ \bottomrule
		\end{tabular}}
	\end{table}
	
	Another common problem that is particularly noted while using 8-bit posit precisions is the vanishing gradient problem -- the gradients become smaller and smaller as the model converges. This is particularly problematic when the model weights are updated with low-precision posits, since they do not have enough resolution for small numbers. As suggested in \cite{Lu2020}, using 16-bit posits for the optimizer and loss is usually enough to allow models to train with low-precision posits. With this observation in mind, this model was trained with a different precision for the optimizer and loss, while using posit(8, 2) everywhere else (see \cref{tab:mixed}). The posit exponent size \gls{es} was fixed at 2, since it gave the best results and simplified the conversion between posits with different \gls{nbits}.
	\begin{table}[t]
		\vspace*{-0.5\baselineskip}
		\centering
		\caption{Accuracy of LeNet-5 trained on Fashion MNIST using posit, quire, and mixed precision. Configuration OxLy means Optimizer (O) with posit(x, 2) and Loss (L) with posit(y, 2), and everything else with posit(8, 2).} 
		\label{tab:mixed}
		\resizebox{\columnwidth}{!}{\begin{tabular}{@{}l|c|cccc@{}}
				\toprule
				Configuration & \textbf{Float} & O12L8 & O12L12 & O12L10 & O10L10 \\ \midrule
				Accuracy [\%] & \textbf{\num{90.42}} & \num{88.40} & \num{90.07} & \num{90.25} & \num{88.08} \\ \bottomrule
		\end{tabular}}
	\end{table}
	
	The obtained results showcase the feasibility of using 8-bit posits, achieving an accuracy very close to 32-bits \acrshort{ieee754}. In particular, while solely computing the optimizer with posit(12, 2) is not enough to achieve a float-like accuracy, when the loss precision is also increased, the model is able to train without any accuracy penalization and using, at most, 12-bit posits. Conversely, if posit(10, 2) is used for both the optimizer and loss, the final accuracy slightly decreases. Therefore, the configuration with 12 bits for optimizer and 10 bits for loss (O12L10 in \cref{tab:mixed}) offers the best compromise in terms of low-precision and overall model accuracy. This configuration will be referred to as posit(8, 2)*, since the loss function and weight update, both computed with higher precision, only represent about \SI{15}{\percent} of the operations that are performed while training the considered models. 
	
\begin{figure}[t]
		\centering
		\includegraphics[width=1\columnwidth]{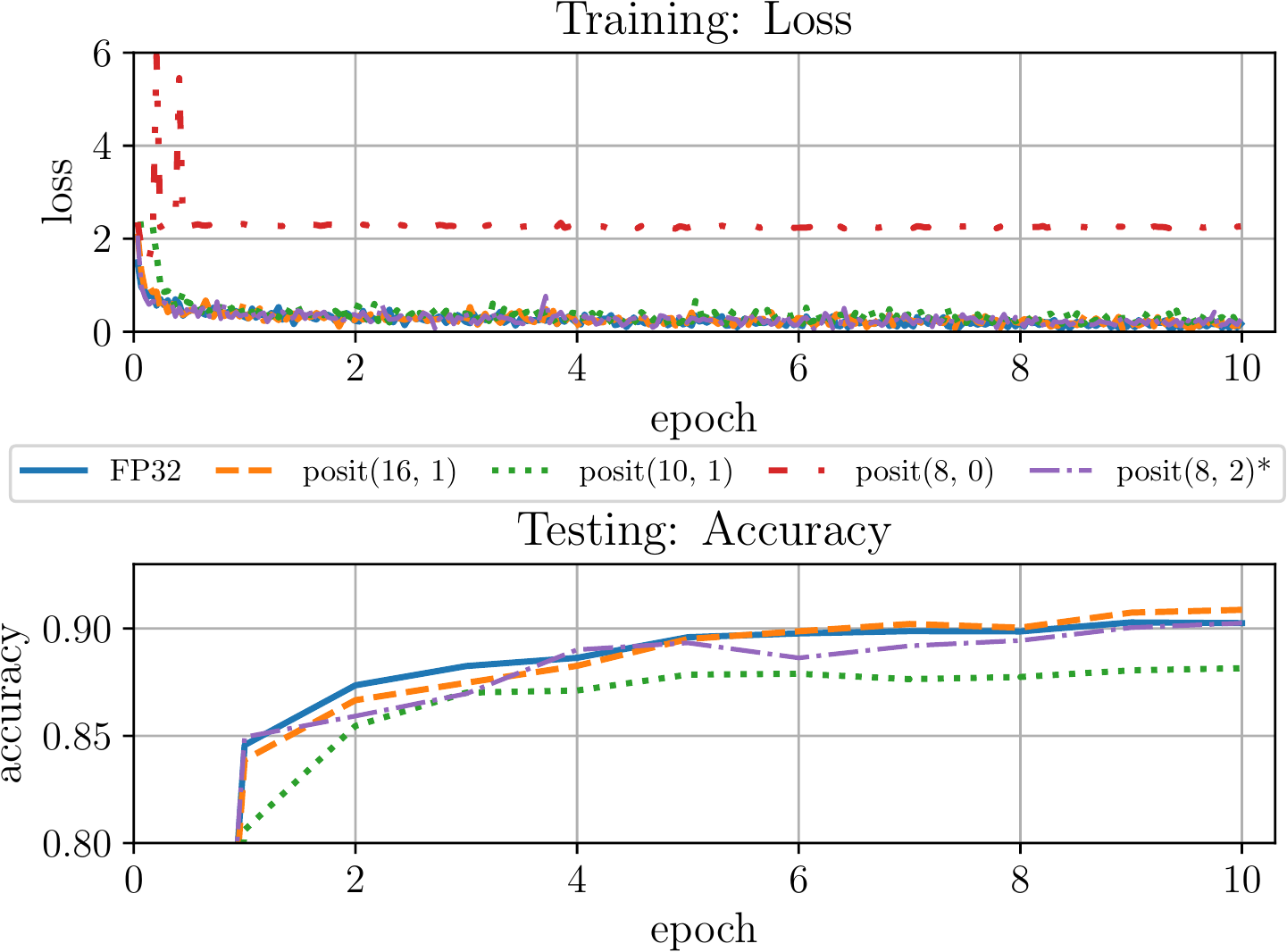}
		\vspace*{-1\baselineskip}
		\caption{Training loss and testing accuracy of LeNet-5 trained on Fashion MNIST using float and different posit precisions. Posit(8, 2)* corresponds to configuration O12L10 of \cref{tab:mixed}.}
		\label{fig:training}
	\end{figure}

	Given the promising results for the Fashion MNIST dataset, the posit(8, 2)* configuration was also used to train LeNet-5 on MNIST and CifarNet on CIFAR-10, validating the proposed mixed configuration. The resulting accuracies are compared against float in \cref{tab:results}. Moreover, a plot of the training progress of LeNet-5 on Fashion MNIST is shown in \cref{fig:training}, comparing different posit configurations and float.
	
	\begin{table}[t]
		\centering
		\caption{Accuracy of \glspl{cnn} trained on MNIST, Fashion MNIST, and CIFAR-10 using float and posit(8, 2)*.}
		\label{tab:results}
		\begin{tabular}{@{}lccc@{}}
			\toprule
			Dataset & MNIST & Fashion MNIST & CIFAR-10 \\
			CNN & LeNet-5 & LeNet-5 & CifarNet \\ \midrule
			Float [\%] & \num{99.19} & \num{90.42} & \num{70.29} \\
			Posit(8, 2)* [\%] & \num{99.17} & \num{90.25} & \num{68.65} \\ \bottomrule
		\end{tabular}
	\end{table}
	
	
	\section{Conclusion}
	
	A new \gls{dnn} framework (PositNN) supporting both training and inference using any posit precision is proposed. The mixed precision feature allows adjusting the posit precision used in each stage of the training network, thus achieving results similar to float. Common \glspl{cnn} were trained with the majority of the operations performed using posit(8, 2) and showed no significant loss of accuracy on datasets such as MNIST, Fashion MNIST, and CIFAR-10.
	
	
	Future work shall make use of this knowledge and framework to devise adaptable hardware implementations of posit units that may exploit this feasibility to implement low-resource and low-power \gls{dnn} implementations while keeping the same model accuracy.
	
	
	
	\section*{Acknowledgments}
	
	Work supported by national funds through Fundação para a Ciência e a Tecnologia (FCT), under the projects\linebreak UIDB/50021/2020 and PTDC/EEI-HAC/30485/2017, and student merit scholarship funded by Fundação Calouste Gulbenkian (FCG).
	
	\vfill\pagebreak
	
	\small
	\bibliographystyle{IEEEbib}
	\bibliography{refs}

\begin{thebibliography}{10}

\bibitem{Thompson2020}
Neil~C. Thompson, Kristjan Greenewald, Keeheon Lee, and Gabriel~F. Manso,
\newblock ``{The Computational Limits of Deep Learning},''
\newblock July 2020,
\newblock arXiv: 2007.05558.

\bibitem{Li2020}
Chuan Li,
\newblock ``{OpenAI}'s {GPT}-3 {Language} {Model}: {A} {Technical}
  {Overview},'' June 2020,
\newblock \url{https://lambdalabs.com/blog/demystifying-gpt-3/}, Accessed on
  2020-10-13.

\bibitem{Schmidhuber2015}
J{\"{u}}rgen Schmidhuber,
\newblock ``Deep learning in neural networks: An overview,''
\newblock {\em Neural Networks}, vol. 61, pp. 85--117, Jan. 2015,
\newblock arXiv: 1404.7828.

\bibitem{IEEE2019}
``{IEEE Standard for Floating-Point Arithmetic},''
\newblock {\em IEEE Std 754-2019 (Revision of IEEE 754-2008)}, pp. 1--84, 2019,
\newblock doi: 10.1109/ieeestd.2019.8766229.

\bibitem{Gustafson2017}
John~L. Gustafson and Isaac Yonemoto,
\newblock ``{Beating Floating Point at its Own Game: Posit Arithmetic},''
\newblock {\em Supercomputing Frontiers and Innovations}, vol. 4, no. 2, pp.
  71--86, June 2017,
\newblock doi: 10.14529/jsfi170206.

\bibitem{Cococcioni2018}
Marco Cococcioni, Emanuele Ruffaldi, and Sergio Saponara,
\newblock ``{Exploiting Posit Arithmetic for Deep Neural Networks in Autonomous
  Driving Applications},''
\newblock in {\em 2018 International Conference of Electrical and Electronic
  Technologies for Automotive}. July 2018, number November, {IEEE},
\newblock doi: 10.23919/eeta.2018.8493233.

\bibitem{Johnson2018}
Jeff Johnson,
\newblock ``{Rethinking floating point for deep learning},''
\newblock Nov. 2018,
\newblock arXiv: 1811.01721.

\bibitem{Langroudi2018}
Seyed Hamed~Fatemi Langroudi, Tej Pandit, and Dhireesha Kudithipudi,
\newblock ``{Deep Learning Inference on Embedded Devices: Fixed-Point vs
  Posit},''
\newblock in {\em 2018 1\textsuperscript{st} Workshop on Energy Efficient
  Machine Learning and Cognitive Computing for Embedded Applications ({EMC}2)}.
  Mar. 2018, pp. 19--23, {IEEE},
\newblock arXiv: 1805.08624, doi: 10.1109/emc2.2018.00012.

\bibitem{Carmichael2019}
Zachariah Carmichael, Hamed~F. Langroudi, Char Khazanov, Jeffrey Lillie,
  John~L. Gustafson, and Dhireesha Kudithipudi,
\newblock ``{Deep Positron: A Deep Neural Network Using the Posit Number
  System},''
\newblock in {\em 2019 Design, Automation {\&} Test in Europe Conference {\&}
  Exhibition ({DATE})}. Mar. 2019, pp. 1421--1426, {IEEE},
\newblock arXiv: 1812.01762, doi: 10.23919/date.2019.8715262.

\bibitem{Carmichael2019a}
Zachariah Carmichael, Hamed~F. Langroudi, Char Khazanov, Jeffrey Lillie,
  John~L. Gustafson, and Dhireesha Kudithipudi,
\newblock ``{Performance-Efficiency Trade-off of Low-Precision Numerical
  Formats in Deep Neural Networks},''
\newblock in {\em Proceedings of the Conference for Next Generation Arithmetic
  2019}, New York, NY, USA, Mar. 2019, vol. Part F1477, pp. 1--9, {ACM},
\newblock arXiv: 1903.10584, doi: 10.1145/3316279.3316282.

\bibitem{Langroudi2019}
Hamed~F. Langroudi, Zachariah Carmichael, John~L. Gustafson, and Dhireesha
  Kudithipudi,
\newblock ``{PositNN Framework: Tapered Precision Deep Learning Inference for
  the Edge},''
\newblock {\em Proceedings - 2019 IEEE Space Computing Conference, SCC 2019},
  pp. 53--59, July 2019,
\newblock doi: 10.1109/spacecomp.2019.00011.

\bibitem{Langroudi2020}
Hamed~F. Langroudi, Vedant Karia, John~L. Gustafson, and Dhireesha Kudithipudi,
\newblock ``{Adaptive Posit: Parameter aware numerical format for deep learning
  inference on the edge},''
\newblock in {\em 2020 {IEEE}/{CVF} Conference on Computer Vision and Pattern
  Recognition Workshops ({CVPRW})}. June 2020, pp. 726--727, {IEEE},
\newblock doi: 10.1109/cvprw50498.2020.00371.

\bibitem{Montero2019}
Ra{\'{u}}l~Murillo Montero, Alberto A.~Del Barrio, and Guillermo Botella,
\newblock ``{Template-Based Posit Multiplication for Training and Inferring in
  Neural Networks},''
\newblock July 2019,
\newblock arXiv: 1907.04091.

\bibitem{Langroudi2019b}
Hamed~F. Langroudi, Zachariah Carmichael, and Dhireesha Kudithipudi,
\newblock ``{Deep Learning Training on the Edge with Low-Precision Posits},''
\newblock July 2019,
\newblock arXiv: 1907.13216v1.

\bibitem{Langroudi2019a}
Hamed~F. Langroudi, Zachariah Carmichael, David Pastuch, and Dhireesha
  Kudithipudi,
\newblock ``{Cheetah: Mixed Low-Precision Hardware \& Software Co-Design
  Framework for DNNs on the Edge},''
\newblock pp. 1--13, Aug. 2019,
\newblock arXiv: 1908.02386.

\bibitem{Lu2019}
Jinming Lu, Siyuan Lu, Zhisheng Wang, Chao Fang, Jun Lin, Zhongfeng Wang, and
  Li~Du,
\newblock ``{Training Deep Neural Networks Using Posit Number System},''
\newblock Sept. 2019,
\newblock arXiv: 1909.03831.

\bibitem{Lu2020}
Jinming Lu, Chao Fang, Mingyang Xu, Jun Lin, and Zhongfeng Wang,
\newblock ``{Evaluations on Deep Neural Networks Training Using Posit Number
  System},''
\newblock {\em {IEEE} Transactions on Computers}, vol. 14, no. 8, pp. 1--1,
  2020,
\newblock doi: 10.1109/tc.2020.2985971.

\bibitem{Murillo2020}
Raul Murillo, Alberto A.~Del Barrio, and Guillermo Botella,
\newblock ``Deep {PeNSieve}: A deep learning framework based on the posit
  number system,''
\newblock {\em Digital Signal Processing}, vol. 102, pp. 102762, jul 2020,
\newblock doi: 10.1016/j.dsp.2020.102762.

\bibitem{Sousa2020}
Leonel Sousa,
\newblock ``Nonconventional computer arithmetic circuits, systems and
  applications,''
\newblock {\em IEEE Circuits and Systems Magazine}, vol. 20, no. 4, pp. 1--26,
  Oct. 2020.

\bibitem{Group2018}
{Posit Working Group},
\newblock ``{Posit Standard Documentation, Release 3.2-draft},'' 2018,
\newblock \url{https://posithub.org/docs/posit_standard.pdf}, Accessed on
  2020-09-24.

\bibitem{Kulisch2012}
Ulrich Kulisch,
\newblock {\em {Computer Arithmetic and Validity}},
\newblock De Gruyter, Berlin, Boston, Jan. 2012,
\newblock doi: 10.1515/9783110301793.

\bibitem{NGATeam2019}
{NGA Team},
\newblock ``{Survey of Posit Hardware and Software Development Efforts},'' Unum
  \& Posit - Next Generation Arithmetic, July 2019,
\newblock \url{https://posithub.org/docs/PDS/PositEffortsSurvey.html}, Accessed
  on 2020-10-16.

\end{thebibliography}
	
\end{document}